\documentclass[lettersize,journal]{IEEEtran}
\usepackage{amsmath,amsfonts}
\usepackage{array}
\usepackage{textcomp}
\usepackage{stfloats}
\usepackage{url}
\usepackage{verbatim}
\usepackage{graphicx}
\usepackage{cite}

\usepackage{subcaption}
\usepackage{multirow}
\usepackage[table,xcdraw]{xcolor}

\usepackage{makecell} 
\usepackage{cite}
\usepackage{tabularx}
\usepackage{booktabs}
\usepackage{caption}

\usepackage{subcaption} 
\usepackage{algorithm}
\usepackage{algorithmic} 

\usepackage{siunitx}   
\usepackage{caption}   

\usepackage{tikz}

\usepackage{pgfplots}
\pgfplotsset{compat=1.18}

\usetikzlibrary{arrows.meta}   

\usetikzlibrary{arrows.meta,shapes,positioning}

\def\BibTeX{{\rm B\kern-.05em{\sc i\kern-.025em b}\kern-.08em
		T\kern-.1667em\lower.7ex\hbox{E}\kern-.125emX}}
\usepackage{balance}
\begin{document}

\title{OASI: Objective-Aware Surrogate Initialization for Multi-Objective Bayesian Optimization in TinyML Keyword Spotting}
	
\author{Soumen Garai, Danilo Pau, \IEEEmembership{Fellow, IEEE} and Suman Samui, \IEEEmembership{Member, IEEE}
\thanks{S. Garai and S. Samui are with the Department of Electronics and Communication Engineering, National Institute of Technology Durgapur, West Bengal 713209, India (e-mail: soumengoroi@gmail.com; ssamui.ece@nitdgp.ac.in). D. Pau is with the System Research and Applications Department, STMicroelectronics, 20864 Agrate Brianza, Italy (e-mail: danilo.pau@st.com).}}

\maketitle

\begin{abstract}
Voice-triggered interfaces rely on keyword spotting (KWS) models that must operate continuously under strict memory, latency, and energy constraints on microcontroller-class hardware. Designing such models therefore requires not only high recognition accuracy but also predictable deployability within limited Flash and SRAM budgets. Bayesian optimization is known to handle accuracy-efficiency trade-offs effectively in multi-objective optimization; however, it is highly sensitive to initialization, particularly in the low-budget regimes of TinyML model optimization.
We propose Objective-Aware Surrogate Initialization (OASI), which seeds surrogate optimization with Pareto-biased solutions generated via multi-objective simulated annealing. Unlike space-filling or heuristic warm-start methods, OASI initializes the surrogate conditioning process with a bias toward feasible accuracy-memory trade-offs, thus avoiding SRAM-violating configurations. OASI improves hypervolume and convergence robustness over Latin hypercube, Sobol, and random initializations under the same budget constraints on a TinyML KWS problem. Hardware-in-the-loop experiments on STM32 microcontrollers verify the existence of deployable and memory-feasible models without incurring extra optimization costs.

\end{abstract}


\section{Introduction} \small{
\IEEEPARstart{S}{peech} assistants like Amazon Echo and Google Home are now seamlessly integrated into our daily lives. Nevertheless, the constant streaming of audio signals to cloud servers raises concerns about network congestion, latency, and privacy \cite{lopez2021deep}. One common remedy for these problems is hybrid processing, where a minimal on-device Keyword Spotting (KWS) component identifies trigger keywords (e.g., “Alexa”) to trigger cloud-based Automatic Speech Recognition (ASR) models. While this helps alleviate communication costs, it comes with very tight constraints on model size (usually below 2 MB) and computational complexity. TinyML \cite{warden2019tinyml} overcomes these challenges by facilitating efficient and privacy-friendly inference on edge devices. In microcontrollers, peak SRAM usage during inference is frequently the primary bottleneck, rather than just the model parameter size, leading to otherwise correct models failing in practice.
	
Many works improve the state-of-the-art in keyword spotting (KWS) through the design of compact models, neural architecture search, or model compression methods like pruning, quantization, and knowledge distillation \cite{garai2024exploring}. While these methods are efficient, they are usually single-objective, providing a single solution. In real-world TinyML KWS, it is necessary to strike a balance between multiple objectives: simultaneously maximize accuracy and minimize memory consumption, latency, and energy costs. Multi-objective Bayesian Optimization (MOBO) is well-positioned for this problem, as it builds surrogate models of accuracy-efficiency trade-offs and seeks Pareto-optimal solutions \cite{garai2025advances}.

The performance of MOBO is very sensitive to initialization, which becomes even more important when the budget of evaluations is limited. The existing literature on structured methods of sampling, such as Latin Hypercube Sampling (LHS), Sobol sequences, and factorial designs, along with meta-learning methods such as MI-SMBO \cite{greif2025structured}, provides early exploration with low overhead. However, objective-agnostic initializations often yield lower-quality surrogate models and slower convergence rates \cite{greif2025structured}. In the TinyML literature, these methods have been used in an ad hoc manner, with little systematic comparison or statistical analysis.
	
To address this, we introduce the Objective-Aware Surrogate Initialization (OASI) method, which leverages multi-objective simulated annealing (MOSA) to generate Pareto-biased seed configurations for multi-objective Bayesian optimization (MOBO) under strict TinyML constraints.
The main contributions of this work are:
\begin{itemize}
		\item We propose OASI, a MOSA-based initialization strategy for MOBO that takes into account the accuracy-model size trade-off in TinyML keyword spotting.
		\item We demonstrate that objective-aware initialization strategies lead to more stable convergence and better-quality Pareto fronts than objective-agnostic strategies such as LHS, Sobol sequences, and random sampling with very limited evaluation budgets.
		\item We demonstrate the efficacy of OASI through hardware-in-the-loop evaluation on STM32 microcontrollers, achieving stable convergence and successful deployment, while objective-agnostic initialization strategies lead to infeasible (out-of-memory) models.
\end{itemize}
\section{Problem Definition and Background}
\subsection{Keyword Spotting Pipeline } 
A continuous audio waveform $x(t)$ is transformed into log-Mel features using short-time Fourier transform (STFT) \cite{lopez2021deep}, yielding $E = \{e_0,\ldots,e_{T-1}\} \in \mathbb{R}^{T \times K}$, where $e_t \in \mathbb{R}^K$. For frame index $i$, a contextual slice is defined as $E[i] = [e_{is-P}, \ldots, e_{is+F}] \in \mathbb{R}^{K \times (P+F+1)}$, with stride $s$, $P$ past, and $F$ future frames. An acoustic model $g_\theta$ outputs softmax posteriors $g_\theta(E[i]) = [p(C_1|E[i]), \ldots, p(C_N|E[i])]^\top$, where $\theta$ is optimized via cross-entropy loss. During streaming inference, temporal smoothing over a window of length $W$ is applied as $\bar{y}_n(t) = \frac{1}{W} \sum_{i=t-W+1}^{t} y_n(i)$, and a keyword is detected when $\bar{y}_{kw}(t) \ge \tau$.
\subsection{Constrained Multi-Objective Optimization for TinyML}
The computational burden and deployment feasibility of the  pipeline are determined primarily by the acoustic model $g_\theta$. Designing $g_\theta$ for TinyML platforms requires selecting architectural hyperparameters $h \in \mathcal{H}$ that jointly maximize predictive accuracy while satisfying strict on-device resource constraints. We therefore formulate model design as a bi-level, black-box constrained multi-objective optimization problem \cite{Loni2020}. The overall bi-level multi-objective optimization workflow is illustrated in Fig. \ref{fig:cmoop_steps}.

Let $\mathcal{D}=\{(x_i,y_i)\}_{i=1}^N$ denote the dataset, partitioned into training ($\mathcal{D}_{tr}$), validation ($\mathcal{D}_{val}$), and test ($\mathcal{D}_{te}$) sets. For a given architectural configuration $h$, the trainable parameters $\theta$ are obtained by solving the lower-level problem
\begin{equation}
	\theta^*(h) \in \arg \min_{\theta} \mathcal{L}(\mathcal{D}_{tr}; \theta, h),
\end{equation}
where $\mathcal{L}$ denotes the supervised loss function. After training, model performance on the validation set is quantified by
\begin{equation}
	Acc(h) = \mathbb{E}_{(x,y)\sim \mathcal{D}_{val}} \left[ \delta(\hat{y}(x;h), y) \right].
\end{equation}

Deployment feasibility on microcontroller-class hardware is governed by non-volatile memory (Flash) and runtime memory (SRAM) requirements. Flash usage is defined as
\begin{equation}
	Flash(h) = \text{bytes}(\theta^*(h), h),
\end{equation}
which accounts for quantized weights and compiled inference code. More critically, peak inference-time SRAM usage is
\begin{equation}
	RAM(h) = \max_t RAM_{act}(h,t),
\end{equation}
capturing activation buffers, operator workspaces, and framework-managed memory arenas. A configuration is deployable only if $RAM(h) \le B_{ram}$, where $B_{ram}$ denotes the available SRAM budget of the target MCU.

We therefore define the constrained bi-objective problem
\begin{equation}
	\min_{h\in\mathcal{H}} \big(f_1(h), f_2(h)\big),
	\quad
	\text{s.t. } RAM(h) \le B_{ram},
\end{equation}
with $f_1(h)=-Acc(h)$ and $f_2(h)=Flash(h)$. Under a fixed evaluation budget $T$, the optimizer seeks a feasible approximation $\mathcal{P}_T$ of the Pareto set $\mathcal{P}^*$.
Although inference latency and cycle count are critical deployment metrics, they are inherently hardware-dependent and require compilation and on-device execution for accurate measurement. Incorporating latency directly into the MOBO objective would require hardware-in-the-loop evaluation at every iteration, significantly increasing the cost of optimization and coupling the search to a specific MCU configuration. Therefore, we consider peak SRAM as a hard feasibility constraint during optimization and evaluate latency and cycle-level efficiency separately in a hardware-in-the-loop validation phase.
\begin{figure}
	\centering
	\includegraphics[scale= 2.0]{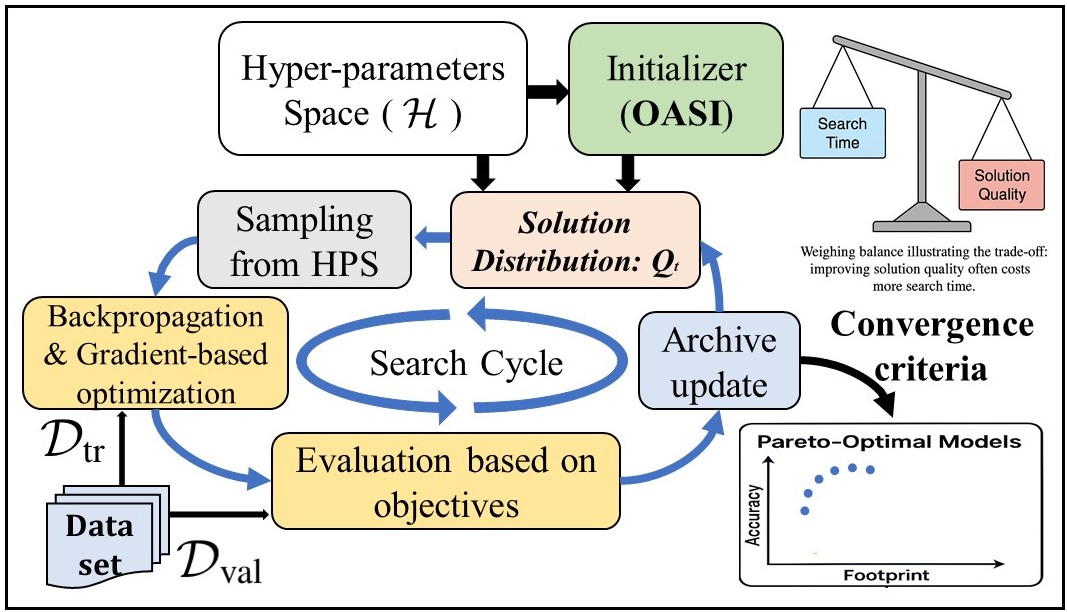}
	\caption{Block diagram of the bi-level multi-objective optimization process: sampling, training, evaluation, archiving, and Pareto front extraction.}
	\label{fig:cmoop_steps}
\end{figure}
\begin{figure}
	\centering
	\includegraphics[scale= 1.2]{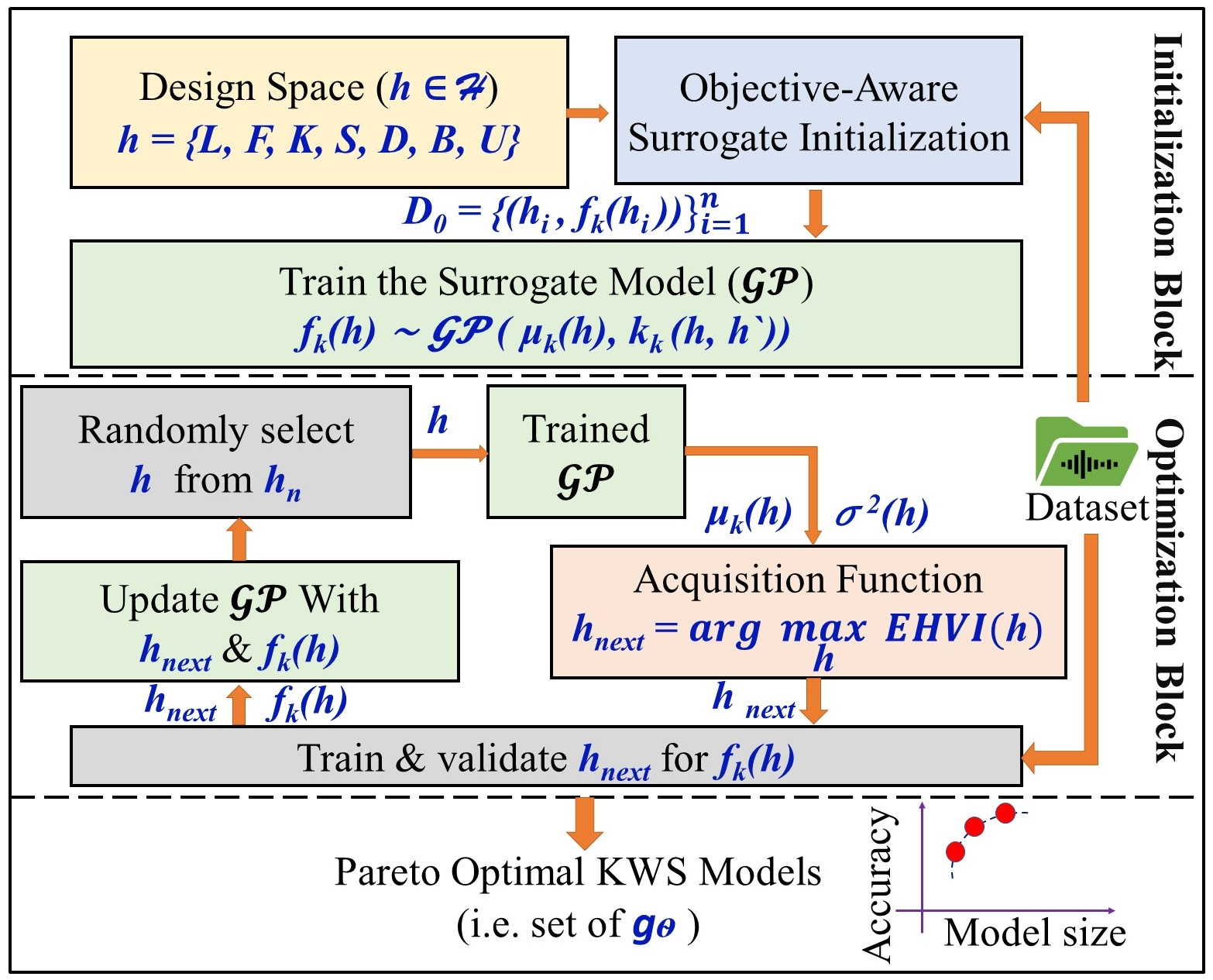} 
	\caption{MOBO framework with proposed OASI framework: MOSA-based initialization with GP-surrogate MOBO for Pareto-optimal KWS models}
	
	\label{fig:block_diagram}
\end{figure}
\section{Proposed OASI for MOBO}
MOBO \cite{daulton2022multi} is a suitable method for dealing with the constrained multi-objective problem inherent in designing TinyML KWS acoustic models. Search is carried out simultaneously to maximize validation accuracy $f_1(h)=\mathrm{Acc}(h)$ and minimize resources such as model size $f_2(h)=\mathrm{Flash}(h)$, with such constraints.

The design vector $h={L,F,K,S,D,B,U}$ encodes architectural hyperparameters, including convolutional depth ($L$), per-layer filters ($F$), kernel sizes ($K$), strides ($S$), dropout rates ($D$), batch normalization flags ($B$), and dense units ($U$). For a given configuration $h$, a model instance $M_h$ is trained with parameters $\theta$ on the training set $\mathcal{D}{\text{tr}}$, yielding optimized weights $\theta^*(h) \in \arg\min{\theta},\mathcal{L}(\mathcal{D}{\text{tr}};\theta,h)$. Because each evaluation requires full training and subsequent validation on $\mathcal{D}{\text{val}}$, the process is computationally expensive. MOBO alleviates this cost by fitting GP surrogates for each objective, $f_k(h)\sim GP(\mu_k(h),k_k(h,h'))$, and proposing new candidates via an acquisition function such as Expected Hypervolume Improvement (EHVI): $h_{\text{next}}=\arg\max_h \alpha_{\text{EHVI}}(h)$.

The quality of the initialization set $D_0={(h_i,f(h_i))}_{i=1}^n$ is critical, as it determines the early posterior means and variances, thereby shaping the exploration trajectory. Space-filling designs such as  LHS or Sobol sequences provide uniform coverage of $\mathcal{H}$ but they overlook the Pareto structure in the objective space.

To address this, we propose OASI, a simulated annealing–based method that generates a Pareto-biased initial archive visualized in Fig. \ref{fig:block_diagram}. OASI employs a MOSA \cite{bandyopadhyay2008simulated}-inspired scheme of short stochastic chains (40–50 iterations). Each chain begins from a random solution $h_{\text{curr}}$ and explores a perturbed neighbor $h_{\text{next}}$, evaluated on accuracy $f_1(h)$ and size $f_2(h)$. Acceptance is probabilistic: $p_{\text{acc}}=1$ if $f_1(h_{\text{next}})>f_1(h_{\text{curr}})$, else $\exp(-(f_1(h_{\text{curr}})-f_1(h_{\text{next}}))/T_{\text{acc}})$; and $p_{\text{size}}=1$ if $f_2(h_{\text{next}})<f_2(h_{\text{curr}})$, else $\exp(-(f_2(h_{\text{next}})-f_2(h_{\text{curr}}))/T_{\text{size}})$. A candidate is accepted only if $u_{\text{acc}}<p_{\text{acc}}$ and $u_{\text{size}}<p_{\text{size}}$, with $u_{\text{acc}},u_{\text{size}}\sim\mathcal{U}(0,1)$. Accepted moves update $h_{\text{curr}}$, while all evaluations are stored in the initializer archive $I_A$.

After all chains complete, $D_0$ is selected from $I_A$ using a maximin rule to ensure broad coverage of $\mathcal{H}$. This objective-aware initialization reduces surrogate bias and improves MOBO’s sample efficiency under TinyML constraints. The complete procedure is summarized in Algorithm~\ref{alg:oasi}.

OASI differs from conventional space-filling or heuristic warm-start methods in that it directly influences surrogate conditioning with Pareto-biased sampling. This helps to improve the calibration of the Gaussian process in the presence of tight TinyML budgets, where violations of SRAM constraints are prevalent. In such low-budget settings, suboptimal surrogate conditioning in the early stages of acquisition may lead to misleading decisions and hinder convergence, which OASI addresses without expanding the budget.
\begin{algorithm}[t]
	\caption{Objective-Aware Surrogate Initialization (OASI)}
	\label{alg:oasi}
	\small
	\begin{algorithmic}[1]
		\REQUIRE Search space $\mathcal{H}$; objective vector $\mathbf{f}(h)$; 
		number of chains $N_{\mathrm{chains}}$; iterations per chain $N_{\mathrm{iter}}$; 
		initial temperatures $T_{\mathrm{acc}}^0$, $T_{\mathrm{size}}^0$; 
		cooling rates $\alpha_{\mathrm{acc}}$, $\alpha_{\mathrm{size}}$; 
		dataset size $n$
		\ENSURE Initial surrogate dataset $D_0$
		\STATE $I_A \gets \emptyset$
		\FOR{$c=1$ \TO $N_{\mathrm{chains}}$}
		\STATE $T_{\mathrm{acc}} \gets T_{\mathrm{acc}}^0$, \; $T_{\mathrm{size}} \gets T_{\mathrm{size}}^0$
		\STATE $h_{\mathrm{curr}} \gets \mathrm{RandomSample}(\mathcal{H})$, \; $\mathbf{f}_{\mathrm{curr}} \gets \mathbf{f}(h_{\mathrm{curr}})$
		\STATE $I_A \gets I_A \cup \{(h_{\mathrm{curr}}, \mathbf{f}_{\mathrm{curr}})\}$
		\FOR{$i=1$ \TO $N_{\mathrm{iter}}$}
		\STATE $h_{\mathrm{next}} \gets \mathrm{Perturb}(h_{\mathrm{curr}})$, \; $\mathbf{f}_{\mathrm{next}} \gets \mathbf{f}(h_{\mathrm{next}})$
		\STATE $I_A \gets I_A \cup \{(h_{\mathrm{next}}, \mathbf{f}_{\mathrm{next}})\}$
		\STATE Compute $p_{\mathrm{acc}}, p_{\mathrm{size}}$; sample $u_1,u_2 \sim \mathcal{U}(0,1)$
		\IF{$u_1 < p_{\mathrm{acc}} \land u_2 < p_{\mathrm{size}}$}
		\STATE $h_{\mathrm{curr}}, \mathbf{f}_{\mathrm{curr}} \gets h_{\mathrm{next}}, \mathbf{f}_{\mathrm{next}}$
		\ENDIF
		\STATE $T_{\mathrm{acc}} \gets \alpha_{\mathrm{acc}} T_{\mathrm{acc}}$, \; $T_{\mathrm{size}} \gets \alpha_{\mathrm{size}} T_{\mathrm{size}}$
		\ENDFOR
		\ENDFOR
		\STATE $D_0 \gets \mathrm{SelectDiverseSubset}(I_A, n)$
		\STATE \textbf{return} $D_0$
	\end{algorithmic}
\end{algorithm}
\begin{figure}
	\centering
	\includegraphics[scale=0.2]{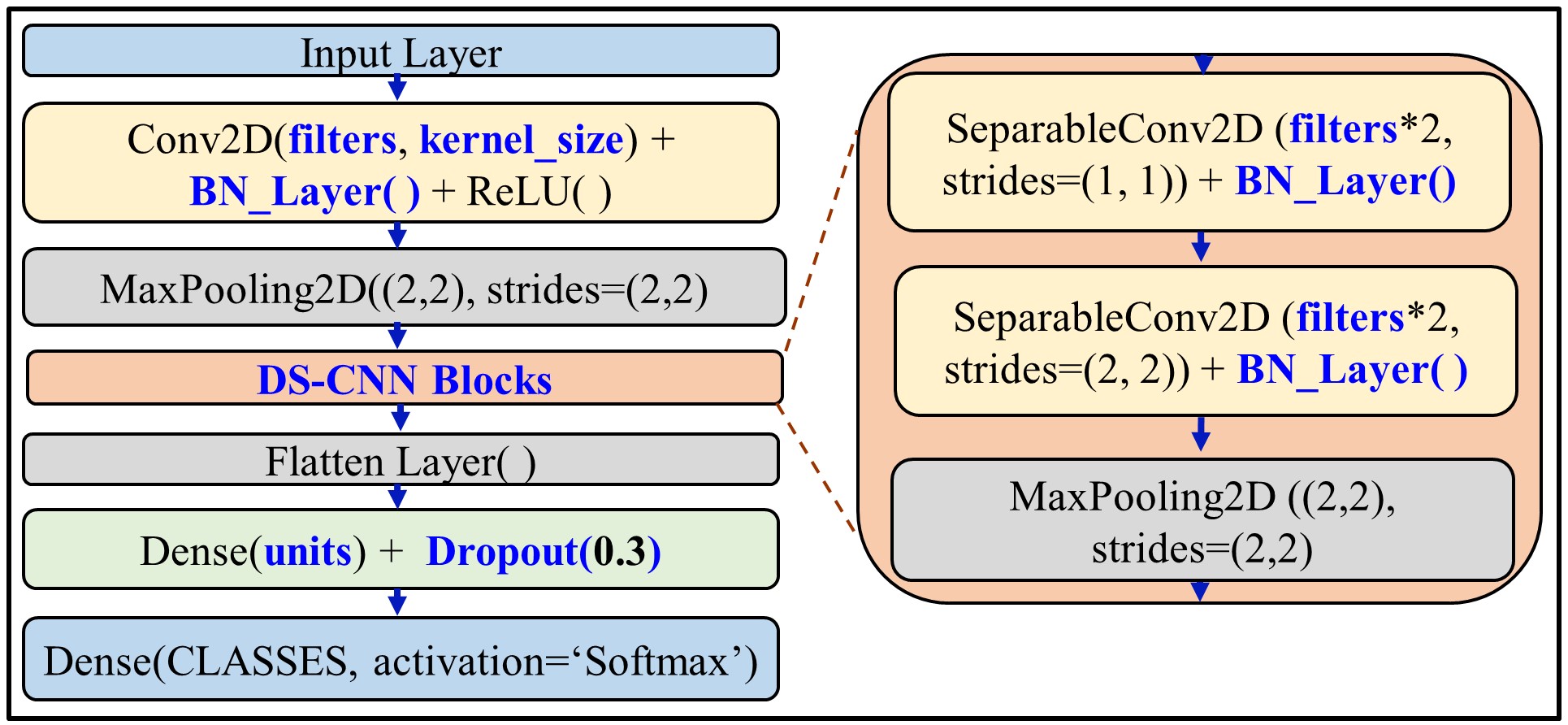}
	\caption{The DSCNN architecture highlighting the key hyperparameters.}
	\label{fig: architecture}
\end{figure}
\section{Experimental result}
We evaluate on Google Speech Commands v2 (GSC) \cite{warden2018speech} with 10 balanced classes (8k/1k/1k train/val/test). Audio is normalized to 1.0\,s at 16\,kHz and mapped to 40-bin log-Mel features (25\,ms window, 10\,ms hop). The backbone is a depthwise separable CNN (DS-CNN) (Fig.~\ref{fig: architecture}), where each standard convolution is factorized into depthwise spatial filtering and pointwise projection, reducing parameters and MACs while preserving accuracy, enabling low-latency TinyML inference. The hyperparameter space is $h \in \{d \in \{1,2,3\},\, w \in [16,64],\, k \in \{3\times3,5\times5\},\, f \in \{1,2,3\},\, \text{BN},\, \text{Dropout}\}$, reflecting embedded constraints. Training uses Adam ($\text{lr}=10^{-3},\beta_1=0.9,\beta_2=0.999$), batch size 64, up to 100 epochs, with early stopping (patience 10). Optimization runs on RTX A4000; deployment metrics are measured exclusively on target MCUs. The complete implementation of OASI and MOBO is publicly available\footnote{\url{https://github.com/sumansamui/OASI-MOBO-TinyML_KWS.git}}.

\begin{table}[t]
	\centering
	\scriptsize
	\caption{Comparison under identical budgets (Top-1 via Tchebycheff; lower $J(h)$ is better).}
	\label{baseline}
	\resizebox{\columnwidth}{!}{%
		\begin{tabular}{lccccc}
			\toprule
			Method & Acc. (\%) & Size (MB) & Iter. Time (s) & Init. Overhead (s) & $J(h)$ \\
			\midrule
			NSGA-II      & 88.9 & 0.210 & 1389.5 & 1499.9 & 0.0127 \\
			MOSA         & 87.9 & 0.114 & 147.5  & 150.0  & 0.0144 \\
			MOBO         & 88.5 & 0.120 & 164.2  & 1501.8 & 0.0145 \\
			\textbf{OASI-MOBO} & \textbf{90.1} & \textbf{0.103} & 164.0 & 1934.9 & \textbf{0.0040} \\
			\bottomrule
	\end{tabular}}
\end{table}
Experiments are divided into (i) optimizer-level comparison and (ii) initialization comparison. In (i), NSGA-II, MOSA, MOBO (random initialization), and OASI-MOBO are evaluated under identical total budget $T$. In (ii), MOBO is fixed and initialization $\in \{\text{LHS},\text{Sobol},\text{Random},\text{OASI}\}$ varies with identical surrogates, acquisition functions, and budgets. Iteration time excludes network training; initialization overhead is reported separately. NSGA-II approximates $\mathcal{P}^*$ via non-dominated sorting ($\mathcal{O}(n^2)$), whereas OASI improves surrogate conditioning through objective-aware seeding.

The comparative outcome shown in Table \ref{baseline} highlights the trade-off between the performance of the proposed OASI framework and the existing multi-objective state-of-the-art approaches \cite{garai2025advances} (NSGA-II, MOBO, and MOSA). To make the comparison fair, all approaches were tested under the same experimental conditions, and the Pareto-optimal solution was obtained using Tchebycheff scalarization (TS): $J(h)=\max\!\left(w_1|f_1(h)-f_1^\star|,\, w_2|f_2(h)-f_2^\star|\right)$ with equal weights and ideal point $(f_1^\star,f_2^\star)$. TS transforms the multi-objective optimization problem into a single minimization problem by combining the normalized accuracy and model size (with equal weights, $w_1 = w_2 = 0.5$) using a weight sum. The approach with the smallest Tchebycheff value is considered the best solution. The best Tchebycheff value is obtained by the proposed OASI approach, which demonstrates its ability to find a 'sweet spot' on the Pareto front that is not found by other approaches. Although NSGA-II is able to fill the Pareto front well, it often faces problems in converging to the boundary points, especially those that are compact and have a reasonable accuracy, due to the stochastic nature of its crossover and mutation operators. In terms of computational complexity, the 'Per Iteration Time' metric measures the time complexity of the search algorithm only, excluding the training time of the model. Among all approaches, NSGA-II is the most computationally expensive, as its computational complexity is dominated by the non-dominated sorting, which grows quadratically with the population size. On the other hand, although OASI is slightly more computationally expensive than MOBO and MOSA, the additional cost is well justified by the significantly better trade-off between accuracy and model size. This is because the objective-oriented initialization strategy of OASI is more effective than the other two approaches.

Fig. \ref{fig:initialization} shows how different initialization methods affect MOBO. LHS, Random, and Sobol spread points across the space but ignore the objectives, so the surrogate often starts with uninformative data. OASI, by contrast, uses MOSA to generate candidates guided by accuracy and size, giving the surrogate a stronger start and leading to better trade-offs under tight budgets. Fig. \ref{fig:progressVstime} illustrates the progression of best validation accuracy and hypervolume under different initialization strategies.

The combined objective score $J(h)$ is shown in Fig. \ref{fig:combined_objective}, where lower values represent better trade-offs between accuracy and size. OASI achieves and maintains the lowest value of J, indicating the best accuracy-size trade-off; Sobol improves incrementally but remains higher than OASI, LHS levels off at a higher value, and Random performs worst in all cases. The plot in Fig. \ref{fig:box} further emphasizes the advantage of OASI by showing a tighter spread and better results compared to the variability of LHS and Random.

\begin{figure}
	\centering
	
	\begin{subfigure}{0.24\textwidth} 
		\includegraphics[width=\textwidth]{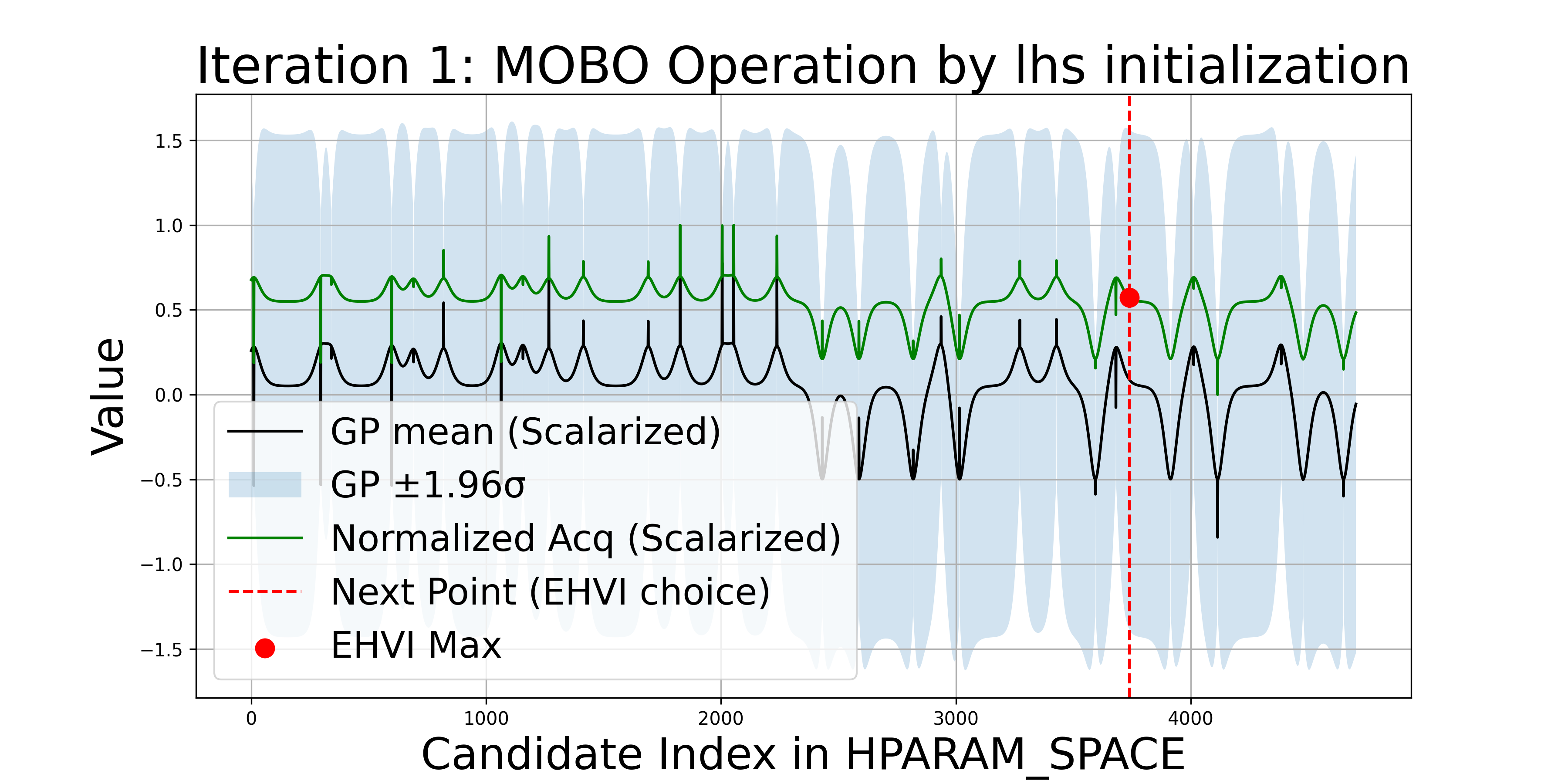}
		\caption{LHS initialization}
		\label{fig:lhs}
	\end{subfigure}
	\begin{subfigure}{0.24\textwidth} 
		\includegraphics[width=\textwidth]{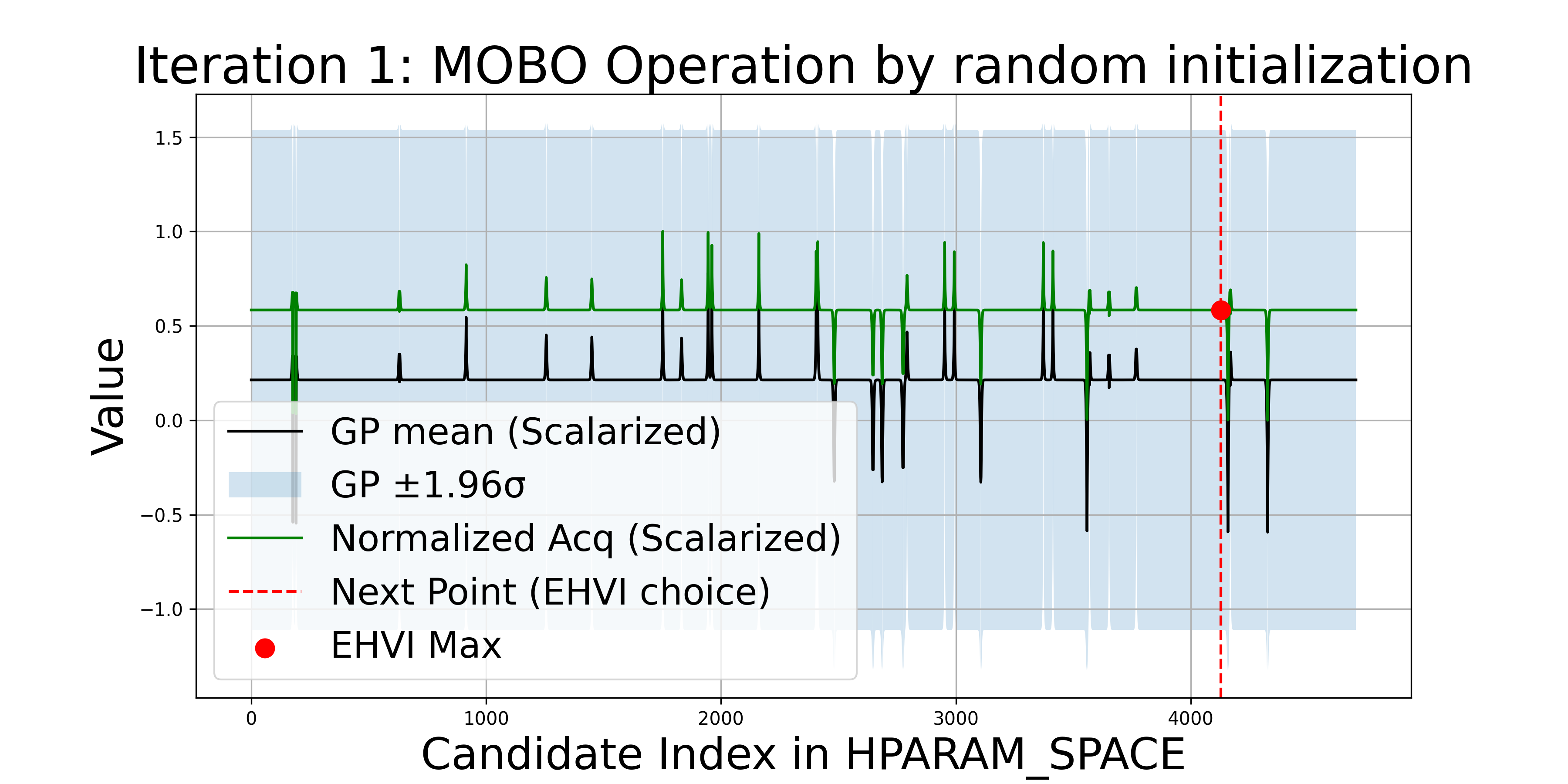}
		\caption{Random initialization}
		\label{fig:random}
	\end{subfigure}
	
	\vspace{0.3cm}
	
	\begin{subfigure}{0.24\textwidth} 
		\includegraphics[width=\textwidth]{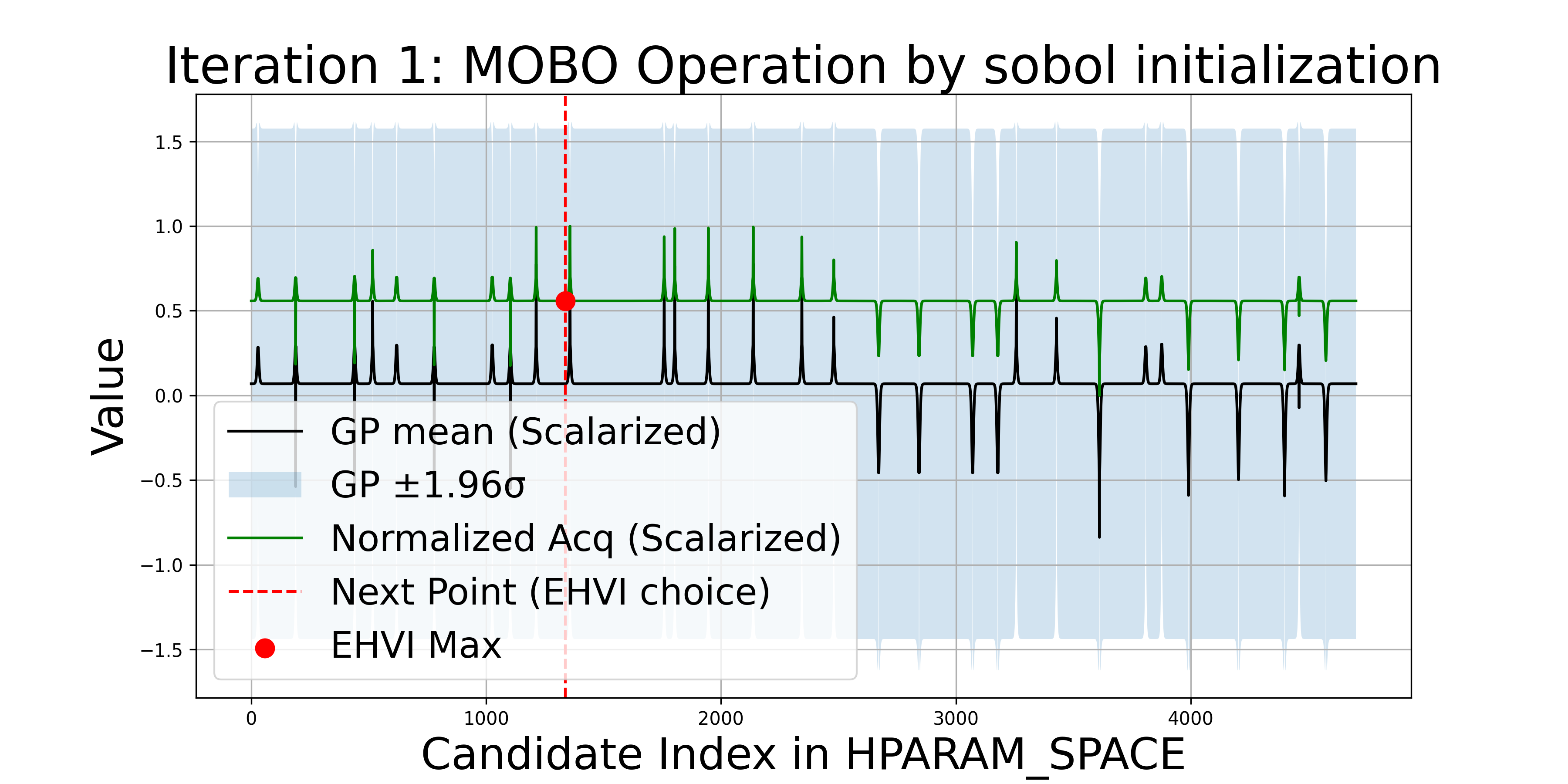}
		\caption{Sobol initialization}
		\label{fig:sobol}
	\end{subfigure}
	\begin{subfigure}{0.24\textwidth} 
		\includegraphics[width=\textwidth]{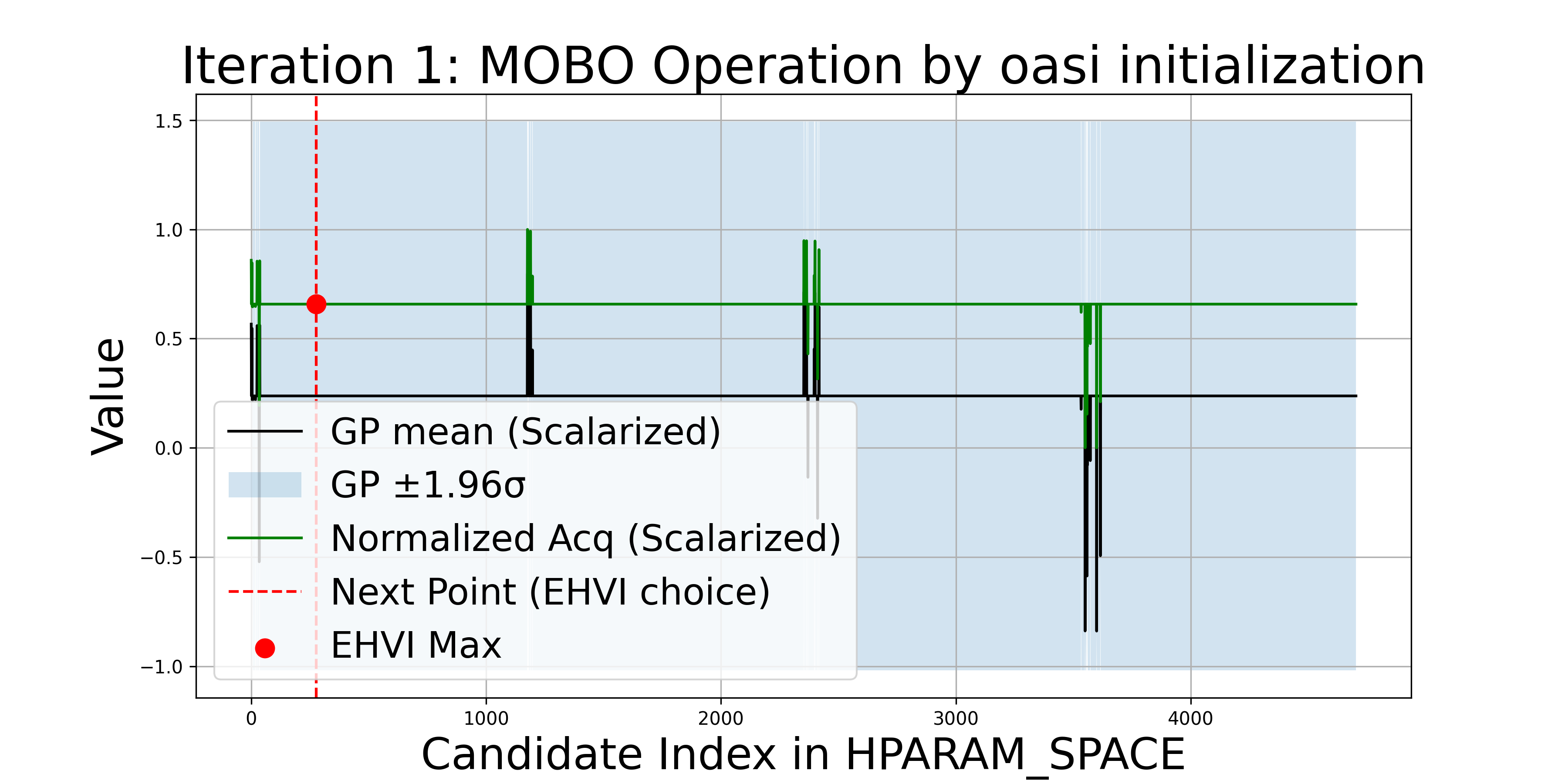}
		\caption{OASI initialization}
		\label{fig:oasi}
	\end{subfigure}
	
	\caption{Effect of initialization strategies on MOBO.}
	\label{fig:initialization}
\end{figure}

\begin{table}
	\centering
	\caption{Performance comparison of initialization strategies}
	\label{tab:performance}
	\begin{tabular}{lccc}
		\toprule
		\textbf{Init.} & \textbf{HV} & \textbf{GD} & \textbf{Time (s)} \\
		\midrule
		LHS    & 0.056336 & 0.462394 & 1501.79 \\
		Random & 0.056773 & 0.005563 & 1501.79 \\
		Sobol  & 0.059255 & 0.003757 & 1562.11 \\
		OASI   & \textbf{0.062748} & \textbf{0.000000} & 1934.88 \\
		\bottomrule
	\end{tabular}
\end{table}

\begin{figure}
	\centering
	
	\begin{subfigure}{0.24\textwidth} 
		\includegraphics[width=\textwidth]{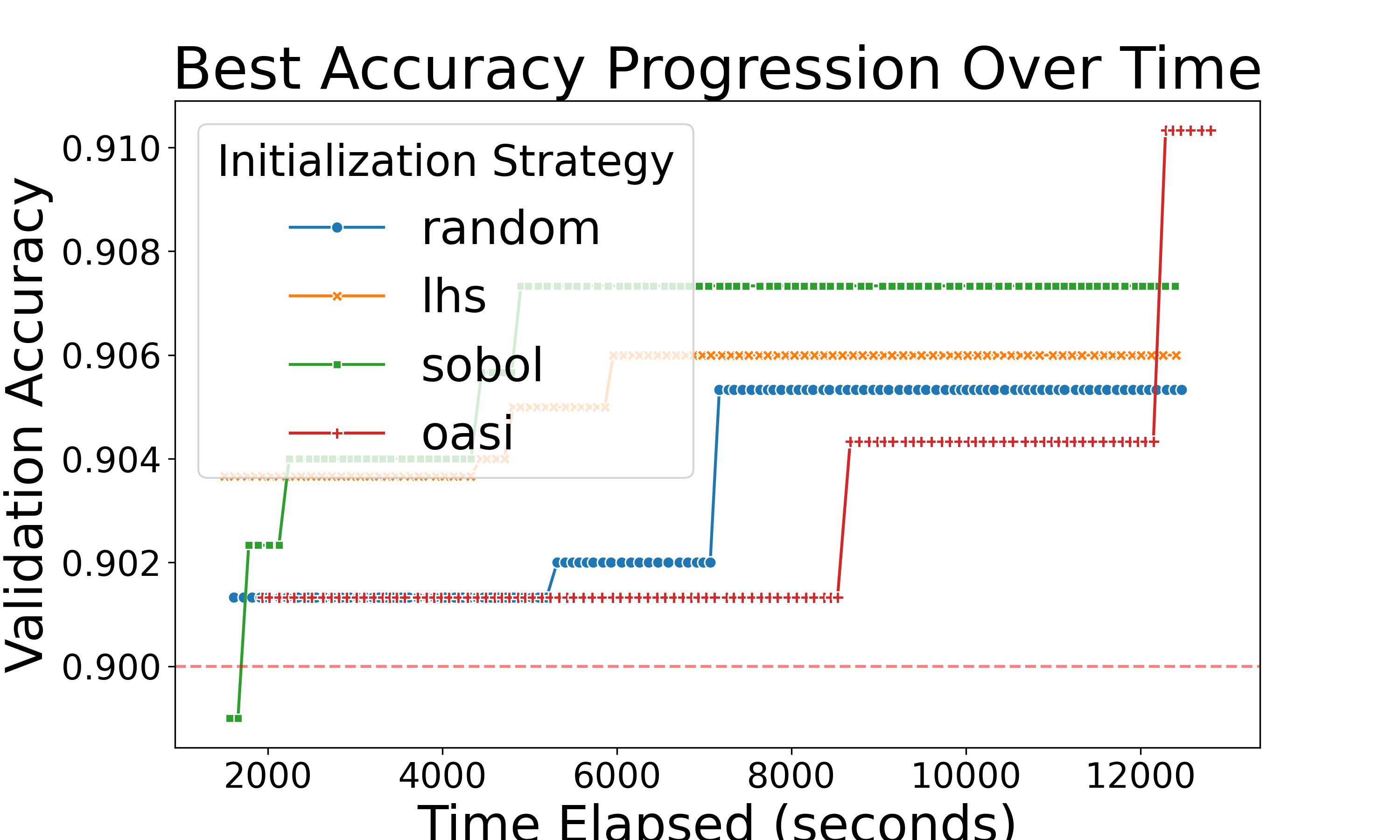}
		\caption{}
		\label{fig:progressVstime-a}
	\end{subfigure}
	\hfill
	\begin{subfigure}{0.24\textwidth} 
		\includegraphics[width=\textwidth]{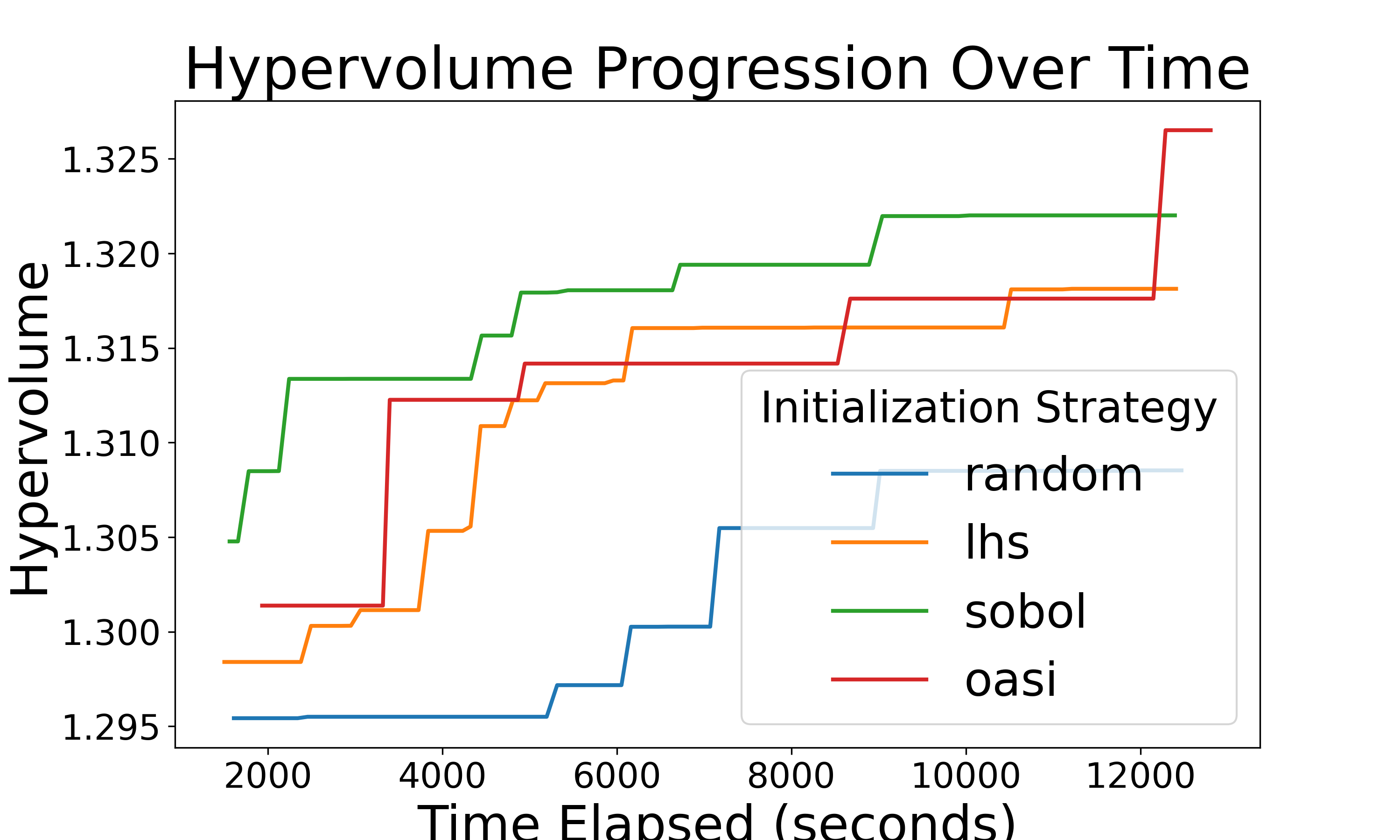}
		\caption{}
		\label{fig:progressVstime-b}
	\end{subfigure}
	
	\caption{Comparison of initialization strategies in terms of (a) best validation accuracy progression and (b) hypervolume progression over time.}
	\label{fig:progressVstime}
\end{figure}

In the bi-objective space $(f_1,f_2)=(-\mathrm{Acc},\mathrm{Flash})$, performance is quantified by hypervolume (HV) and generational distance (GD). HV measures dominated objective space relative to a reference point; GD measures convergence to the reference Pareto front. Table \ref{tab:performance} compares initialization methods by hypervolume (HV), generational distance (GD), and runtime. OASI yields the highest HV and lowest GD (zero), demonstrating superior convergence to the Pareto front, with only marginally higher runtime than others. 

Objective-agnostic initializations (LHS, Sobol, Random) provide uniform coverage but weaker early surrogate conditioning. OASI seeds via MOSA-guided sampling, improving early Pareto proximity. OASI exhibits superior HV progression and sustained reduction in $J(h)$, while Sobol plateaus despite competitive early behavior. Reduced inter-run variance indicates improved convergence stability. 

Statistical evaluation using the Kruskal--Wallis test ($H=5.40$, $p=0.144$, $\eta^2=0.0007$) does not reach conventional significance; however, OASI uniquely achieves $\mathrm{GD}=0$ with consistently lower dispersion, indicating improved convergence stability despite modest statistical separation. 

\begin{figure}
	\centering
	
	\begin{subfigure}{0.24\textwidth} 
		\includegraphics[width=\textwidth]{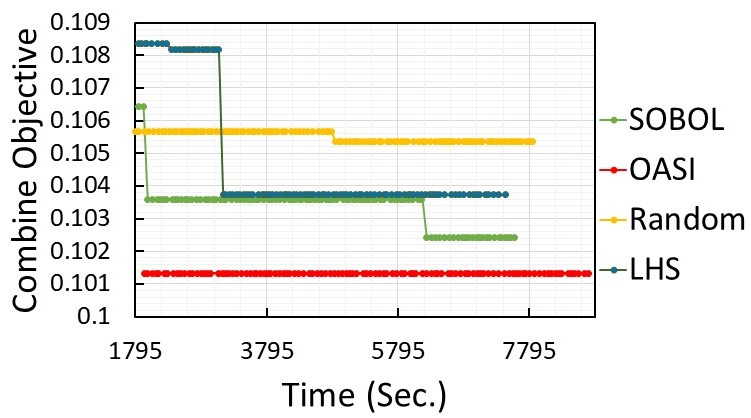}
		\caption{}
		\label{fig:combined_objective}
	\end{subfigure}
	\hfill
	\begin{subfigure}{0.24\textwidth} 
		\includegraphics[width=\textwidth]{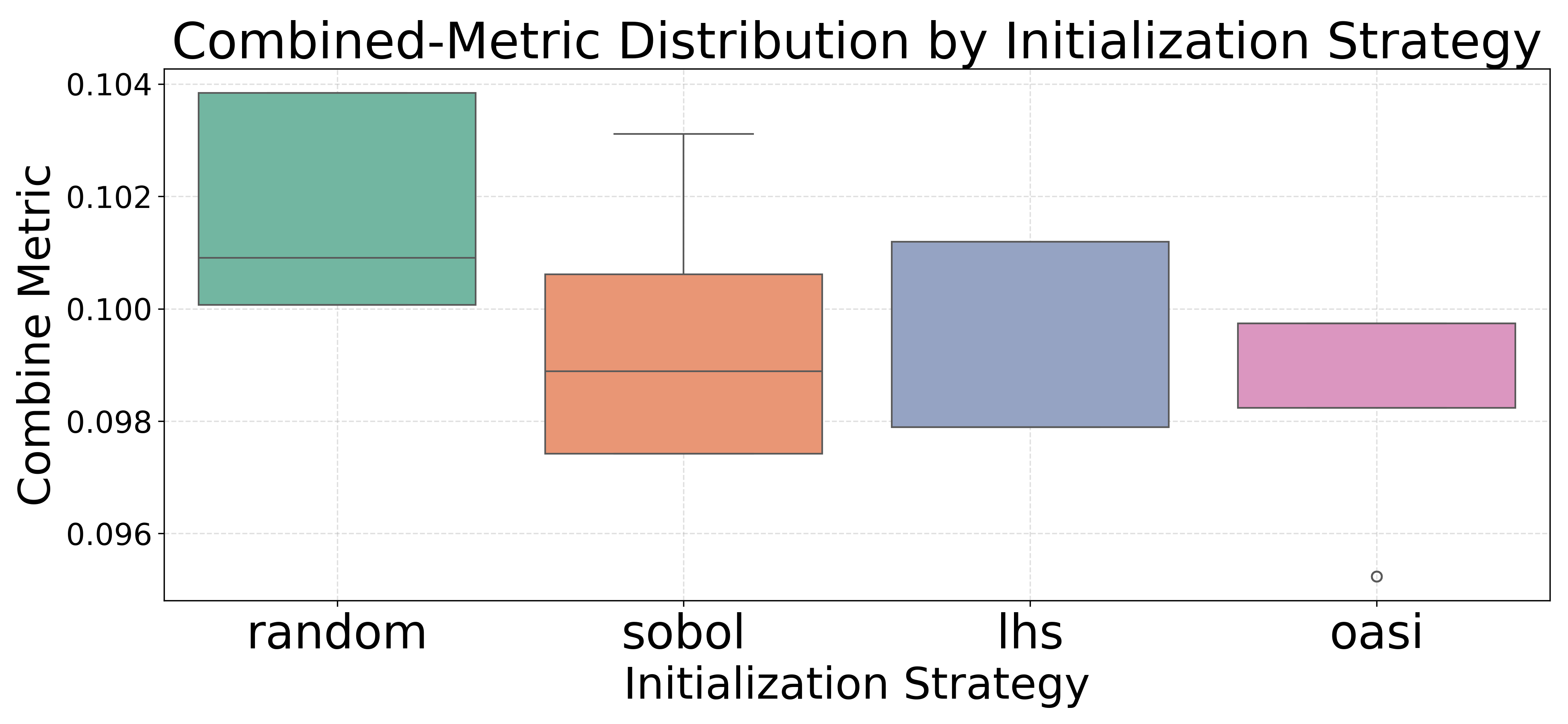}
		\caption{}
		\label{fig:box}
	\end{subfigure}
	
	\caption{(a) Combined objective over time for different initialization strategies and (b) Distribution of the combined objective across different initialization strategies.}
	
\end{figure}

\begin{figure}
	\centering
	\includegraphics[width=0.4\textwidth]{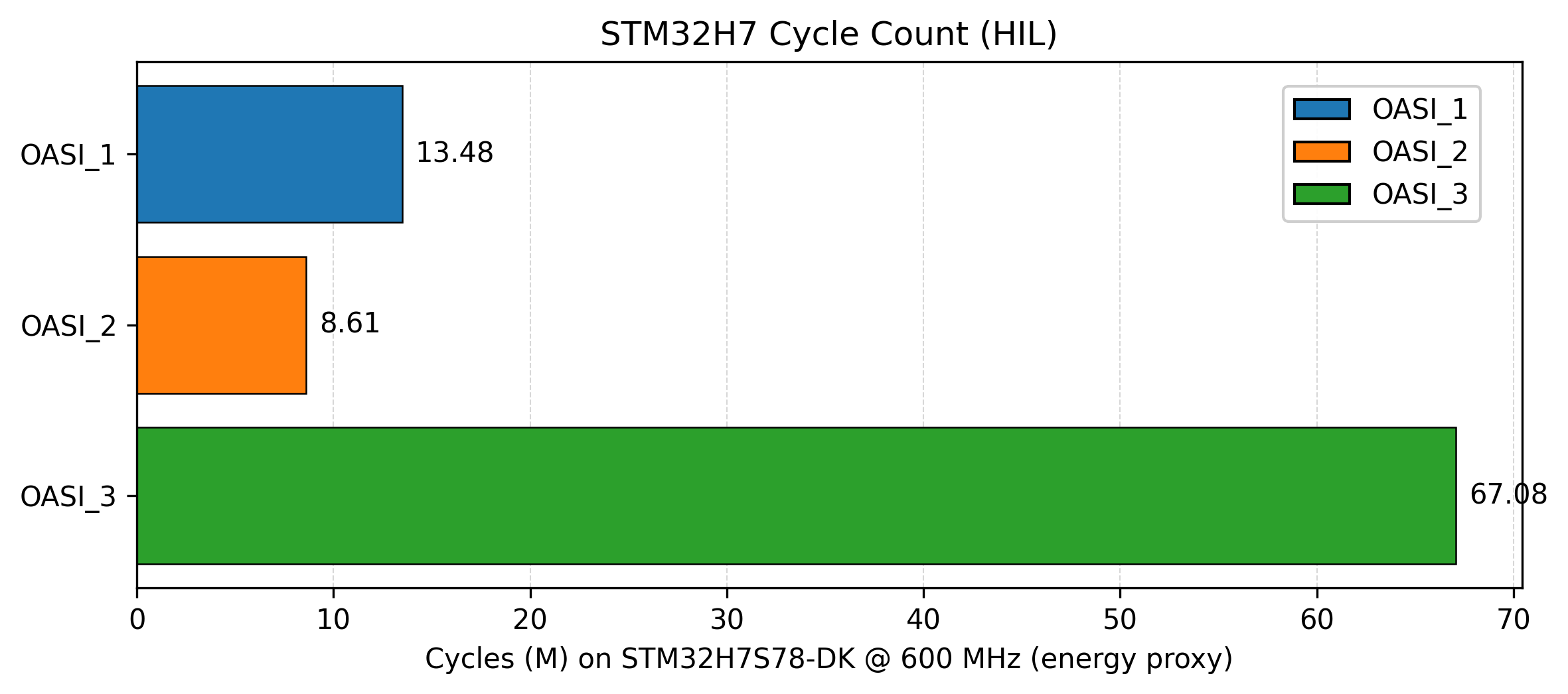}
	\caption{HIL measured cycle counts on STM32H7S78-DK (600\,MHz) for the top three OASI-selected models.}
	\label{fig:h7_cycles_bar}
\end{figure}

\begin{table*}
	\centering
	\caption{HIL results and board-wise DI of the top three OASI models on STM32 platforms}
	\label{tab:hil}
	\footnotesize
	\setlength{\tabcolsep}{3.5pt}
	\begin{tabular}{l c c c c c c c c c c c}
		\toprule
		\multirow{2}{*}{\textbf{Model}}
		& \textbf{Acc.} & \textbf{Params} & \textbf{Flash} & \textbf{Peak RAM} & \textbf{MACC}
		& \multicolumn{3}{c}{\textbf{Latency (ms)}}
		& \multicolumn{3}{c}{\textbf{DI}} \\
		\cmidrule(lr){7-9}\cmidrule(lr){10-12}
		& \textbf{(\%)} & \textbf{(K)} & \textbf{(KB)} & \textbf{(KB)} & \textbf{(M)}
		& \makecell{\textbf{H7}\\\textbf{(600\,MHz)}}
		& \makecell{\textbf{F469}\\\textbf{(180\,MHz)}}
		& \makecell{\textbf{F401}\\\textbf{(84\,MHz)}}
		& \makecell{\textbf{H7}}
		& \makecell{\textbf{F469}}
		& \makecell{\textbf{F401}} \\
		\midrule
		OASI\_1 & 90.1 & 29.88 & 73.2  & 53.5  & 6.8  & 22.5  & 117.1 & 231.1 & 0.831 & 0.428 & 0.029 \\
		OASI\_2 & 88.7 & 29.51 & 62.8  & 51.0  & 6.8  & 14.3  & 84.3  & 170.0 & \textbf{0.865} & \textbf{0.541} & \textbf{0.132} \\
		OASI\_3 & 91.0 & 71.69 & 117.7 & 135.5 & 47.1 & 111.8 & 548.3 & OOM   & 0.432 & 0.000 & -- \\
		\bottomrule
	\end{tabular}
	
	\vspace{2mm}
	\footnotesize{
		\center{OOM denotes out-of-memory due to insufficient SRAM for the activation arena. }
		
	}
\end{table*}
\textbf{Hardware-in-the-Loop Deployment:}
To assess practical deployability, the int8-quantized models were compiled using STM32Cube.AI and executed on physical STM32 microcontrollers via the ST Edge AI Developer Cloud\footnote{\url{https://stedgeai-dc.st.com}}. This hardware-in-the-loop (HIL) framework evaluates optimized inference binaries on real MCUs, capturing memory access behavior and kernel scheduling effects not reflected by theoretical FLOPs or MACC (multiply–accumulate operations) \cite{Zhuang2024}. The evaluated platforms include STM32H7, STM32F469, and STM32F401 microcontrollers (abbreviated as H7, F469, and F401 in Table III).
Table~\ref{tab:hil} summarizes cross-board deployment results. OASI\_2 requires only 51.0~KB peak RAM and 62.8~KB Flash, enabling deployment on the SRAM-constrained NUCLEO-F401RE, whereas OASI\_3 exceeds low-tier SRAM limits (135.5~KB) and results in OOM failure. Although OASI\_3 achieves the highest accuracy (91.0\%), its latency scales poorly on CPU-based MCUs (111.8~ms on STM32H7, 548.3~ms on STM32F469). In contrast, OASI\_2 achieves 0.315~ms latency on STM32N6570-DK with 95.6\% operator offloading to the Neural-ART accelerator, demonstrating hardware portability.
Fig.~\ref{fig:h7_cycles_bar} further shows that OASI\_2 attains the lowest cycle count on STM32H7. Despite identical MACC (6.8\,M) for OASI\_1 and OASI\_2, measured cycles and latency differ significantly, confirming that memory reuse and kernel scheduling dominate effective MCU execution cost. The near-linear relationship between latency and cycle count indicates compute-bound behavior under sufficient memory conditions.
These observations motivate a hardware-aware Deployability Index (DI), defined as
\begin{equation}
	DI_{i,h} =
	\left(\prod_{u \in \{R,F,T\}}
	\left(1 - \frac{U_{i,h}^{(u)}}{B_h^{(u)}}\right)\right)
	\mathbb{I}\!\left[
	\frac{U_{i,h}^{(u)}}{B_h^{(u)}} \le 1 \;\; \forall u
	\right],
	\label{eq:DI}
\end{equation}
where $U_{i,h}^{(R)}$, $U_{i,h}^{(F)}$, and $U_{i,h}^{(T)}$ denote peak SRAM, Flash usage, and latency, respectively, and $B_h^{(R)}, B_h^{(F)}, B_h^{(T)}$ represent hardware or application limits. DI measures deployment slack relative to resource bounds; infeasible configurations yield $DI=0$. The multiplicative formulation penalizes saturation in any single resource dimension, reflecting the strict bottleneck behavior of MCU deployment. The DI values in Table~\ref{tab:hil} confirm that OASI\_2 consistently occupies the most favorable deployment region across MCU tiers. While OASI\_3 offers marginal accuracy gains, its elevated memory and runtime cost substantially reduce deployment slack, particularly for devices with hard resource constraints.

\section{Conclusion}
This brief introduces OASI, an objective-aware initialization approach for multi-objective Bayesian optimization in TinyML keyword spotting. OASI improves the convergence stability of Pareto-biased seeds over space-filling and heuristic warm-start approaches, always achieving zero generational distance with equal budgets. Hardware-in-the-loop experiments on STM32 MCUs verify that the models chosen by OASI meet strict SRAM and Flash requirements without compromising competitive accuracy-efficiency tradeoffs. The Deployability Index, a quantitative deployment readiness metric,  further positions objective-aware initialization as a viable solution for hardware-feasible TinyML optimization.


%

\bibliographystyle{IEEEtran}
\bibliography{ref}

@inproceedings{garai2024exploring,
  title={Exploring tinyml frameworks for small-footprint keyword spotting: A concise overview},
  author={Garai, Soumen and Samui, Suman},
  booktitle={2024 International Conference on Signal Processing and Communications (SPCOM)},
  pages={1--5},
  year={2024},
  organization={IEEE}
}

@article{greif2025structured,
  title={Structured sampling strategies in Bayesian optimization: evaluation in mathematical and real-world scenarios},
  author={Greif, Lucas and H{\"u}bschle, Niklas and Kimmig, Andreas and Kreuzwieser, Simon and Martenne, Anatole and Ovtcharova, Jivka},
  journal={Journal of Intelligent Manufacturing},
  pages={1--31},
  year={2025},
  publisher={Springer}
}

@article{garai2025advances,
  title={Advances in Small-Footprint Keyword Spotting: A Comprehensive Review of Efficient Models and Algorithms},
  author={Garai, Soumen and Samui, Suman},
  journal={arXiv preprint arXiv:2506.11169},
  year={2025}
}

@inproceedings{daulton2022multi,
  title={Multi-objective bayesian optimization over high-dimensional search spaces},
  author={Daulton, Samuel and Eriksson, David and Balandat, Maximilian and Bakshy, Eytan},
  booktitle={Uncertainty in Artificial Intelligence},
  pages={507--517},
  year={2022},
  organization={PMLR}
}

@article{bandyopadhyay2008simulated,
  title={A simulated annealing-based multiobjective optimization algorithm: AMOSA},
  author={Bandyopadhyay, Sanghamitra and Saha, Sriparna and Maulik, Ujjwal and Deb, Kalyanmoy},
  journal={IEEE transactions on evolutionary computation},
  volume={12},
  number={3},
  pages={269--283},
  year={2008},
  publisher={IEEE}
}

@article{warden2018speech,
  title={Speech commands: A dataset for limited-vocabulary speech recognition},
  author={Warden, Pete},
  journal={arXiv preprint arXiv:1804.03209},
  year={2018}
}

@article{lopez2021deep,
  title={Deep spoken keyword spotting: An overview},
  author={L{\'o}pez-Espejo, Iv{\'a}n and Tan, Zheng-Hua and Hansen, John HL and Jensen, Jesper},
  journal={IEEE Access},
  volume={10},
  pages={4169--4199},
  year={2021},
  publisher={IEEE}
}

@book{warden2019tinyml,
  title={Tinyml: Machine learning with tensorflow lite on arduino and ultra-low-power microcontrollers},
  author={Warden, Pete and Situnayake, Daniel},
  year={2019},
  publisher={O'Reilly Media}
}

@InProceedings{Zhuang2024,
  author       = {Zhuang, Brenda and Pau, Danilo},
  booktitle    = {2024 IEEE International Conference on Consumer Electronics (ICCE)},
  title        = {A practical framework for designing and deploying tiny deep neural networks on microcontrollers},
  year         = {2024},
  organization = {IEEE},
  pages        = {1--6},
}

@Article{Loni2020,
  author    = {Loni, Mohammad and Sinaei, Sima and Zoljodi, Ali and Daneshtalab, Masoud and Sj{\"o}din, Mikael},
  journal   = {Microprocessors and Microsystems},
  title     = {DeepMaker: A multi-objective optimization framework for deep neural networks in embedded systems},
  year      = {2020},
  pages     = {102989},
  volume    = {73},
  publisher = {Elsevier},
}

}

\end{document}